# Exploring Embeddings for Measuring Text Relatedness: Unveiling Sentiments and Relationships in Online Comments

Anthony Olakangil
*Bellarmine College Preparatory*
San Jose, USA
A.Olakangil26@bcp.org

Cindy Wang
*Barrington High School*
Barrington, USA
cindy.wang.student@gmail.com

Justin Nguyen
*Bellarmine College Preparatory*
San Jose, USA
Justin.Nguyen24@bcp.org

Qunbo Zhou
*Los Osos High School*
Rancho Cucamonga, USA
Bob.z123006@gmail.com

Kaavya Jethwa
*California High School*
San Ramon, USA
jethwa.kaavya@gmail.com

Jason Li
*William Fremd High School*
Palatine, USA
Jason.ye.li.7@gmail.com

Aryan Narendra
*Basis Independent Silicon Valley*
San Jose, USA
aryan.narendra1@gmail.com

Nishk Patel
*University of Illinois*
Urbana-Champaign, USA
nishkdp2@illinois.edu

Arjun Rajaram
*University of Maryland*
College Park, USA
arajara1@terpmail.umd.edu

*Abstract*—After the COVID-19 pandemic caused internet usage to grow by 70%, there has been an increased number of people all across the world using social media. Applications like Twitter, Meta Threads, YouTube, and Reddit have become increasingly pervasive, leaving almost no digital space where public opinion is not expressed. This paper investigates sentiment and semantic relationships among comments across various social media platforms, as well as discusses the importance of shared opinions across these different media platforms, using word embeddings to analyze components in sentences and documents. It allows researchers, politicians, and business representatives to trace a path of shared sentiment among users across the world. This research paper presents multiple approaches that measure the relatedness of text extracted from user comments on these popular online platforms. By leveraging embeddings, which capture semantic relationships between words and help analyze sentiments across the web, we can uncover connections regarding public opinion as a whole. The study utilizes pre-existing datasets from YouTube, Reddit, Twitter, and more. We made use of popular natural language processing models like Bidirectional Encoder Representations from Transformers (BERT) to analyze sentiments and explore relationships between comment embeddings. Additionally, we aim to utilize clustering and Kl-divergence to find semantic relationships within these comment embeddings across various social media platforms. Our analysis will enable a deeper understanding of the interconnectedness of online comments and will investigate the notion of the internet functioning as a large, interconnected brain.

## I. INTRODUCTION

The aftermath of the 2020 COVID-19 pandemic created a newfound surge of online platforms and user-generated content that revolutionized interaction, with comments being pivotal for reflecting opinions. Because of this, the relevance of social media increased, which sparked interest in this research. Public sentiment offers insight into the general populace's standings on certain topics. By using Sentiment analysis (SA) which involves classifying text as containing positive, negative, or neutral sentiment, overall sentiment and much more can be accessed. Natural Language Processing (NLP) has advanced in the field of text embeddings [1], which entails encoding text into vectors to capture word relationships [2]. This study uses models like BERT [3] and methods like dimensionality reduction and clustering to identify global trends. Kl-Divergence measures semantic and sentiment convergence or divergence over time, another method we employed.

Previous works in the field of NLP have explored methods like dimensionality reduction and Canonical Correlation Analysis to understand relationships between different vectors [4-5]. However, these approaches fall short of revealing a collective consensus or divergence of opinions across more than one user. In contrast, our research aims to extend this analysis to explore collective consensus or divergence of opinions on various topics.

Employing methods like K-means clustering and systems like ADRMine for pharmacovigilance and medical purposes also falls under the vast topic of NLP [6]. Techniques like the bag of character n-grams have also been utilized to identify relationships, such as detecting hate speech and toxic content on Twitter [7]. Embeddings have been used to detect deepfake images and videos as well [8].

Neri et al. [9] utilized 1000 Facebook posts and examined individual comment's opinions. This methodology provided insights into public opinions toward products. While these studies have provided an understanding of individual sentiments, they fall short in measuring collective sentiment accurately, as well as using multiple platforms. This is an obstacle because certain platforms may contain a concentrated demographic of people who hold biases, meaning deriving public sentiment from just one platform will not give a comprehensive understanding of global, shared opinion.

Our research, on the other hand, aims to analyze relationships between multiple comments and compare sentiments and semantics across multiple platforms on a larger scale. The field of opinion mining has exploited deep learning strategies to delve into the semantics of social media [10]. These approaches use models like Multi-Layered Bi-Directional Long-Short-Term Memory (MBiLSTM) [11] to achieve high accuracy when classifying comments. While these approaches are closely related to our proposal, their goals stop at classification.

Other studies have introduced novel approaches to sentiment analysis, like Balahur's study of analyzing sentiment in Twitter data with minimal linguistic processing [12], as well as Wang and Li analyzing sentiment within photos [13]. However, our study focuses on analyzing sentiment strictly within text and uncovering more nuanced patterns. Additionally, studies analyzing sentiment in specific tweets [14] or utilizing the Skip-gram model used by Asr, Zinkov, and Jones primarily aim to optimize a model or use strictly SA [15], not find more patterns. While using the Skip-gram model has been proven to find asymmetrical relationships with high accuracies, parameterization is elaborate and complex [16]. Our focus extends beyond optimizing word embedding algorithms, deriving an expanded spectrum of patterns through multiple models.

Extracting meaningful insights from social media comments poses a challenge due to their unstructured nature. Because of this, we propose a novel study addressing key challenges, distinguishing us from prior works. By analyzing sentiment across platforms, we transcend SA limitations. This approach captures the interconnections and attitudes among online users. Moving beyond sentiment classification, our study underlying relationships via clustering and Kl-Divergence. This research offers a more complete understanding of sentiment dynamics. This paper is organized as follows: Section II covers tested models, their working diagrams, datasets used, and trend analysis metrics. Section III discusses accuracy metrics, visualizes the model results, and expresses key takeaways. Section IV concludes the paper, and discusses options for building further on this topic.

## II. METHODS

### 2.1 Data Collection

Data for this project was sourced from YouTube, Reddit, Twitter, and Amazon. Various contexts were analyzed, exploring cross-platform internet dynamics, and mitigating potential biases.

We used the "Trending YouTube Video Statistics and Comments" dataset [17], containing views, likes, category, and comments for top daily YouTube videos. Another YouTube dataset from Kaggle contained over 270,000 comments [18]. To diversify our analysis, we incorporated the Sentiment140 dataset, with 1,600,000 tweets and sentiment labels, which are useful for prediction models [19]. For a broader scope, 5 distinct Subreddit pages represented topics like news, gaming, movies, the NFL, and relationships [20]. Finally, our testing encompassed an Amazon Product Reviews dataset, a compendium of more than 568,000 consumer reviews [21]. Our combined dataset contained over 2.6 million data points.

Among these datasets, there were 4 different platforms. However, except for the Amazon reviews, there were two different datasets for each platform, totaling up to 7 datasets. Fairly equal data distribution among platforms ensured an accurate, comprehensive view of patterns. In various tests, we utilized this dataset, not singular platforms.

#### 2.1.1 Dataset Cleaning

Comments on social media contain noise, irregularities, and format variations, necessitating rigorous data cleaning. Emojis were removed, hashtags were dropped, and URLs were replaced. All letters were set to lowercase due to their unhelpful and potentially distracting nature during SA. Initial word position affects how the model treats it, but punctuation removal distorts the original meaning.

### 2.2 Embeddings

Embedding models were used to project text into a multi-dimensional space. Differing from previous research, we used more than one NLP model to garner more information, gain a more complete understanding of the interconnections of the internet, and investigate which models performed best for our specific tasks.

After evaluating various embedding models, we investigated both word and sentence embeddings to better understand the text at two different levels of granularity.

#### 2.2.1 BERT

One of the embedding models we used in our analysis was BERT (Bidirectional Encoder Representations from Transformers), an NLP model developed by Google. BERT is unique in its use of bidirectional context analysis, with a novel approach of Masked Language Modeling (MLM) that allows BERT to predict missing words in a sentence [3]. Figure 1 shows a pipeline of BERT's embedding process.

#### 2.2.2 GloVe

Another embedding model tested in our experiments is the GloVe (Global Vectors) model, an unsupervised learning algorithm for developing word embeddings created by Stanford University. It derives semantic relationships using both local and global statistics.

#### 2.2.3 ALBERT

A Lite BERT, named ALBERT, utilizes architecture similar to BERT. However, we discovered it performs much better than BERT due to its capability to generate superior SA accuracies. It is also less computationally intensive, utilizing cross-layer parameter sharing.

#### 2.2.4 RoBERTa

Created by Facebook AI, RoBERTa (a Robustly Optimized BERT Training Approach), utilizes a similar ar-

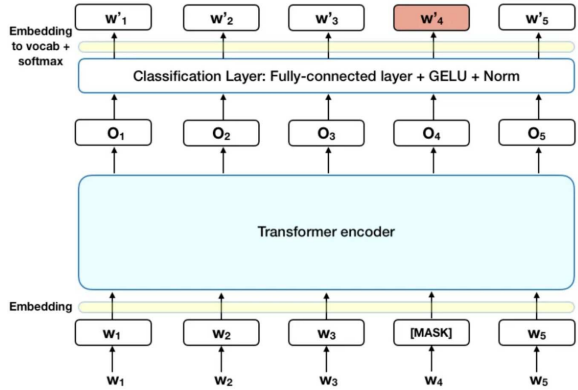

Fig. 1. BERT Model. Source: Adapted from [22]

architecture to BERT. When initially pre-trained, researchers conducted a more extensive hyper-parameter optimization search than what was done with BERT. We looked into 3 checkpoints of RoBERTa including sentiment-RoBERTa-large-english (SiEBERT), Sn-Xlm-RoBERTa-Base-Snli-Mnli-Anli-Xnli (XLM-RoBERTa), and Ko-sRoBERTa-Multitask, and found varying results.

2.2.5 Instructor-xl

Instructor-xl is an embedding model that generates text embeddings for specific domains and tasks by using task-related text instructions. It extracts relevant information from its pre-trained data to create detailed embeddings based on domain-specific jargon, all without requiring fine-tuning or hyper-parameter adjustments. However, when creating clusters, convenience for quality was not a good trade-off. It was computationally expensive, and the clusters generated were of poor quality.

2.2.6 All Mpnet Base

All Mpnet Base is another embedding model that maps both sentences and paragraphs into a vector space. All Mpnet Base can determine sentence similarity, which aligns with our goals of finding relationships and patterns between comments across various social media platforms.

2.2.7 ELMo

This word embedding model, similar to GloVe, is called ELMo (Embeddings from Language Models). It uses a deep, bi-directional LSTM model to create word representations, which allows ELMo to generate different embeddings for the same word used in different contexts.

2.2.8 FastText

The last model we tested was the FastText model, chosen because it generated embeddings much quicker than other models. However, it was found the model could not classify well, with relatively low accuracies for SA.

2.3 Relationship Measurements

To determine the relationship across comments, we utilized many techniques, including standard cosine similarity. Once the words had been embedded, the following equation could be used to determine the similarity between texts, where a similarity of 1 indicates a strong similarity and -1 indicates opposite meanings.

$$\cos \theta = \frac{\mathbf{A} \cdot \mathbf{B}}{\|\mathbf{A}\| \cdot \|\mathbf{B}\|} = \frac{\sum_{i=1}^{n} A_i \cdot B_i}{\sqrt{\sum_{i=1}^{n} A_i^2} \cdot \sqrt{\sum_{i=1}^{n} B_i^2}} \quad (1)$$

Equation 1 demonstrates how to determine the cosine similarity between two vectors, A and B, and $n$ is the length of the vectors. $i$ is the index variable used to sum over the $n$ elements of the vectors.

Other methods we used to measure relations include clustering, for which we used K-means, Principal Component Analysis (PCA), weighted averages, and Kl-Divergence. PCA is a valuable technique for extracting features. Because of our large dataset, PCA was useful for retaining only the most important features in a vector, which can expedite training times and reduce noise. As seen in Figure 2, the plateau represents a point of diminishing returns.

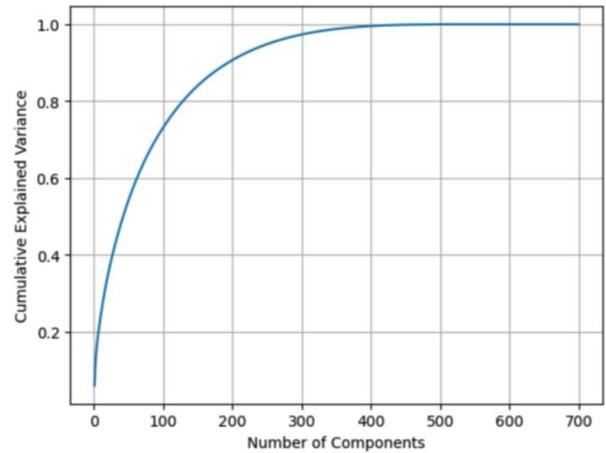

Fig. 2. Principal Component Analysis (PCA) Graph for Feature Extraction, a Preprocessing Method

Clustering was utilized to place each comment in a specific category or "cluster". To determine how many clusters to create, we used the elbow method to find the optimal value of K. The y-axis, "Inertia", indicates the sum of the squared distances of data to an assigned cluster centroid. The x-axis indicates the optimal number of clusters, as indicated in Figure 3. The point the graph starts to plateau or form an "elbow" indicates the most optimal number of n_clusters for the data. We measured the quality of the clusters with the silhouette scoring metric, where a score of -1 means objects are closer to neighboring clusters than to the centroid of their assigned cluster, 0 meaning there are overlapping clusters, and a score close to +1 indicates high quality, distinct clusters.

Finally, we measured the relative entropy, or Kl-Divergence of two probability distributions. A divergence of 0 means the two distributions are identical, and 1 indicates unique probability distributions. Experiments involving Kl-Divergence revealed themes in terms of sentiment and semantics across specific time periods. The formula for KL-Divergence is shown in Equation 2, where $P(i)$ and $Q(i)$ are the probabilities of the ith category according to the distributions of P and Q, respectively.

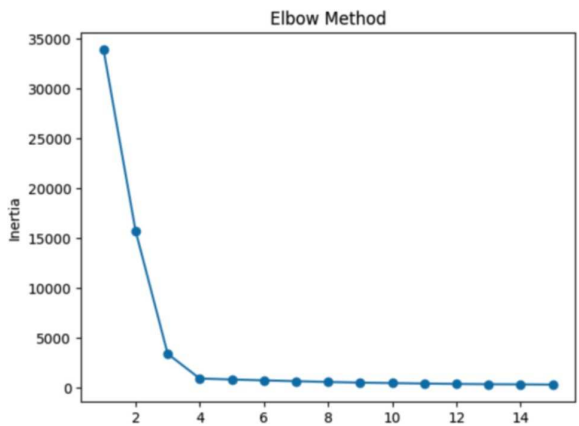

Fig. 3. Elbow Method Graph for Ko_RoBERTa_Multitask

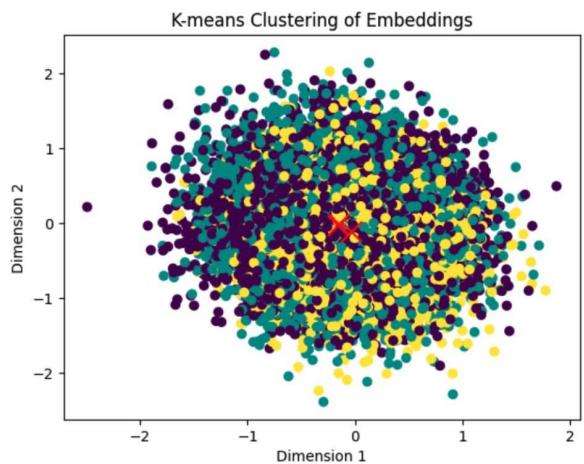

Fig. 4. XLM-RoBERTa K-means Clustering Graph

$$KL(P\|Q) = \Sigma(P(i) * \log(P(i)/Q(i))) \qquad (2)$$

### III. RESULTS

#### 3.1 Initial Embedding Model Testing

We tested a variety of embedding models to determine which would best fit our needs, eventually narrowing our choice down to one. Most of these required no finetuning. We were able to compare the performance of the various models across multiple different tasks such as clustering, Kl-Divergence, and sentiment analysis. All tests were run with the mixed-platform dataset.

#### 3.1.1 RoBERTa

We investigated three different RoBERTa-based models, finding two had quite similar results. The first model was Ko_RoBERTa_Mulitask, a Hugging Face model based on RoBERTa's architecture. We tested this model on multiple different tasks, including clustering and sentiment analysis. In SA, we found its accuracy to be 74.28% when tested on the mixed platform dataset. When clustering, the model achieved the lowest silhouette score of all the models, 0.027.

The next RoBERTa-based model was called XLM-RoBERTa. This model achieved 77.36% accuracy when tested for sentiment analysis and a 0.14 silhouette score for its clusters.

Figure 4 is a visualization for K-means clustering of embeddings using the XLM-RoBERTa model. The clusters lacked any type of structure or definition.

The final RoBERTa-based model tested was sentiment-RoBERTa-large-english (SiEBERT). The model performed optimally with a 99.37% accuracy when tested on a mixed dataset for SA, as well as a 0.83 silhouette score for clustering.

#### 3.1.2 Instructor-xl

Multiple tests were run on Instructor-xl to find trends, but the model was rendered useless in comparison to others. Despite its convenience with the domain feature, the quality of the embeddings it generated was subpar. Primarily aiming to leverage its clustering feature, we achieved a silhouette score of 0.066. Instructor-xl struggled with complex language and thus created many overlapping clusters.

#### 3.1.3 ALBERT

We trained this model on multiple datasets. Given its vector embeddings were more detailed than Instructor-xl, it was able to generate better clusters, with a silhouette score of 0.24. However, it was still relatively low compared to SiEBERT.

#### 3.1.4 FastText

We trained Facebook's FastText model on the mixed-platform dataset, achieving only 63.29% accuracy. This is likely due to outdated context measurement and insufficient pre-training. Embedding quality directly affects SA accuracy, and here, it was inadequate. Because of this, we decided not to pursue it any further in terms of other experiments, like clustering.

#### 3.1.5 Comparison of Results

TABLE I
ANALYSIS AND ACCURACY OF EACH MODEL ON DIFFERENT TASKS

| Model | SA Accuracy |
|---|---|
| SiEBERT | 99.37% |
| ALBERT | 87.20% |
| XLM-RoBERTa | 77.36% |
| Ko-sRoBERTa-Multitask | 74.28% |
| FastText | 63.29% |

| Model | Silhouette Score |
|---|---|
| SiEBERT | 0.83 |
| ALBERT | 0.24 |
| XLM-RoBERTa | 0.14 |
| Instructor-xl | 0.066 |
| Ko-sRoBERTa-Multitask | 0.027 |

The final model selection was SiEBERT. After numerous other tests, SiEBERT proved to have the best performance across all methods, namely categorical Kl-Divergence, K-means clustering, and sentiment classification.

## 3.2 Further Analysis with SiEBERT

### 3.2.1 SA

Once SA was run on multiple individual-platform datasets, as well as the mixed dataset containing content from 4 different platforms, the general sentiment was determined for each to help better understand the clustering assignments. In each instance, the weighted average of sentiment classifications was taken, with an average of -1 being fully negative, and +1 being fully positive. We found everything to be largely balanced, as most only tipped slightly towards a negative sentiment. SiEBERT's sentiment prediction distributions were approximately 40% positive and 60% within the mixed dataset. This shows there is an almost equal distribution of sentiment in the data, with an edge toward negative sentiment. The following table contains the weighted averages found in different datasets.

TABLE II
WEIGHTED AVERAGES TO FIND OVERALL SENTIMENT

| Platform | Weighted Average |
|---|---|
| Mixed | +0.59 |
| YouTube | +0.12 |
| Reddit | -0.17 |
| YouTube | -0.19 |
| Reddit | -0.22 |
| Twitter | -0.24 |

### 3.2.2 Clustering

The Elbow Method was used to conveniently find the optimal number of clusters, rather than manual hyper-parameter optimization. The number of clusters generated reveals how the comments varied, but given only 4-6 clusters were needed for our mixed dataset, it was understood semantics and the overall attitude within each cluster were strongly correlated.

The main experiment run using this model was clustering random comments from a mixed-platform dataset that only generated clusters if the comments met a certain similarity threshold, 0.8. This ensured the comments within a cluster revolved around the same object, entity, or theme. By running these tests, the magnitude of shared sentiment can be better understood. These tests were successful, generating clusters with a consistent silhouette score of 0.83. As seen in this visualization, the clusters generated were a lot more distinct than in previous models, illustrated in Figure 5. Because there were multiple, well-separated clusters, it was made clear the internet may not be highly interconnected. This may be true within each cluster, as thoughts and semantics are related, but as a whole, viewpoints strongly differ regarding similar topics.

### 3.2.3 KL Divergence

The second main test we ran with SiEBERT was categorical Kl-Divergence. Using an Amazon review dataset with more than 500,000 reviews, we looked at 2 time periods, 2004-2008, and 2009-2012, and a divergence of 0.87 was determined. This showed the way reviews were written, the sentiment displayed by the text, and the selection of words drastically changed over time, as a Kl-Divergence near 1.0 indicates very different probability distributions. Given this information, we concluded the internet is not very consistent and is easily influenced.

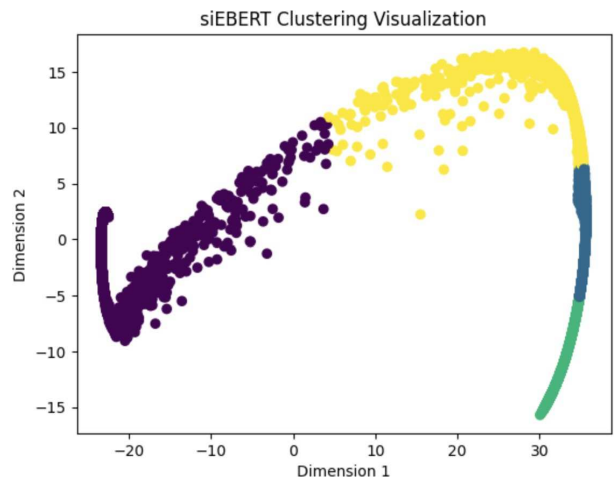

Fig. 5. SiEBERT Clustering Graph

### 3.2.4 Additional Analysis

Various other tests were run for further data analysis, like calculating the general sentiment. This can be used in favor of politicians and businessmen. Using a dataset from Twitter, we found the overall sentiment towards Prime Minister of India Narendra Modi was -0.2, where an average of -1 meant everything classified as negative. However, acknowledging the possible usage of bots would skew results, this sentiment may not be entirely true, as bots can also impact the opinion to a degree. However, based on these results, this shows there is not an overwhelming favor to one side or the other, and the opinions expressed are equally divided.

TABLE III
BEST RESULTS WITH IDENTIFYING TRENDS

| Model | Platform | Metrics |
|---|---|---|
| SiEBERT | Mixed | SA: 99.37% |
| | | Silhouette Score: 0.83 |
| | | Kl-Divergence: 0.87 |

## IV. CONCLUSION

This paper explores the significance of word embeddings in analyzing components of major platforms. While evaluating various embedding models, we narrowed down our model selection to RoBERTa-based models, which showed the most promising results with an average SA accuracy of 85%. Most notably, SiEBERT excelled in SA and classified with an accuracy of 99.37%. Internet opinions appear evenly split, evidence being the sentiment classification distributions. The use of Kl-Divergence highlighted the semantic deviation in Amazon reviews over 12 years, indicating large changes in the way user reviews were written and expressed over time. Clustering online comments with SiEBERT obtained a silhouette score of 0.83. SiEBERT revealed distinct clusters but also diverse internet viewpoints. Several other experiments were conducted, specifically focusing on finding attitudes toward public figures, like the Prime Minister of India, Narendra Modi. Our findings revealed a slightly negative public opinion towards Modi as of 2015, which, if used in further studies with newer data, could provide valuable information for politicians on how they are perceived on various platforms. However, the usage of bots and automated responses may have skewed the results. This notion applies to all of our experiments, where bots could most definitely impact the results found. Our experiments on our data (although a relatively small representation of the internet) all point to the notion that the internet is not that interconnected after all, and the data analyzed had a relatively equal balance of positive and negative sentiment.

In the future, an option for improvement of this research involves creating new data with a way to filter out bots or pre-generated responses to mitigate the influence of bots. This would help to find the genuine opinions of the people. Possible topics for exploration include extending the principles to explore sentiment variations across global regions, which could be used to analyze political rifts and tensions. Additionally, expanding to more niche social media platforms would allow for a broader look at the internet as a whole. Incorporating data from platforms catering to specific demographics could reveal biases and nuanced patterns. This includes region-exclusive and age-targeted platforms. Expanding this project's focus, we could analyze specific comment sections of individual videos, studying divergence or convergence trends over time. An intriguing aspect to explore would be to determine if top comments shape the tone of subsequent comments over time. Furthermore, studies could delve deeper into SA within specific industries, particularly marketing and politics. This would open doors for more practical applications of SA findings.

*A. Acknowledgements*

We would like to thank all those who have assisted us throughout the research process, especially the BLAST AI research program.

Finally, we would like to express our gratitude to the researchers whose previous work has served as an inspiration and foundation for this study. We would also like to acknowledge Kaggle and the authors of the Kaggle datasets we used.